\pdfoutput=1
\documentclass[11pt]{article}

\usepackage[preprint]{latex/acl}

\usepackage{times}
\usepackage{latexsym}
\usepackage[T1]{fontenc}
\usepackage[utf8]{inputenc}
\usepackage{microtype}
\usepackage{inconsolata}
\usepackage{graphicx}

\usepackage[most]{tcolorbox}
\usepackage{multirow}
\usepackage{arydshln}  % for dashed rowlines
\usepackage{enumitem}
\usepackage{xurl}

\tcbset{
  mygraybox/.style={
    enhanced,
    colback=gray!20,
    colframe=gray!80,
    boxrule=0.5mm,
    arc=4mm,
    auto outer arc,
    width=\linewidth,
    fonttitle=\bfseries,
    coltitle=black,
    attach boxed title to top center={yshift=-2mm},
    boxed title style={
      colback=gray!80,
      colframe=gray!80,
      sharp corners,
      size=small
    },
    top=6mm,         % Add vertical space between top border and content
    bottom=2mm,      % Optional: adjust bottom spacing too
    left=2mm,        % Optional: adjust left/right padding
    right=2mm,
    boxsep=4pt       % Space between frame and content
  }
}

\title{\textsc{LuxInstruct}: A Cross-Lingual Instruction Tuning Dataset For Luxembourgish}

\author{Fred Philippy\textsuperscript{1} \And
        Laura Bernardy\textsuperscript{1} \And
        Siwen Guo\textsuperscript{2}
        \AND
         \And Jacques Klein\textsuperscript{1} \And \\ \\
         \textsuperscript{1}SnT, University of Luxembourg, Luxembourg \vspace{0.1cm} \\
         \textsuperscript{2}Luxembourg Institute of Science and Technology, Luxembourg \And
         Tegawendé F. Bissyandé\textsuperscript{1} \And
         }

\begin{document}
\maketitle

\begin{abstract}
Instruction tuning has become a key technique for enhancing the performance of large language models, enabling them to better follow human prompts. However, low-resource languages such as Luxembourgish face severe limitations due to the lack of high-quality instruction datasets. Traditional reliance on machine translation often introduces semantic misalignment and cultural inaccuracies. In this work, we address these challenges by creating a cross-lingual instruction tuning dataset for Luxembourgish, without resorting to machine-generated translations into it. Instead, by leveraging aligned data from English, French, and German, we build a high-quality dataset that preserves linguistic and cultural nuances. 
We provide evidence that cross-lingual instruction tuning not only improves representational alignment across languages but also the model's generative capabilities in Luxembourgish. This highlights how cross-lingual data curation can avoid the common pitfalls of machine-translated data and directly benefit low-resource language development.\footnote{Code and data are provided at \url{https://github.com/fredxlpy/LuxInstruct}.}
\end{abstract}
\section{Introduction}
Instruction tuning improves the ability of Large Language Models (LLMs) to follow user prompts by fine-tuning them on curated instruction–output pairs, leading to better generalization and alignment with human intent \citep{ouyang2022traininglanguagemodelsfollow}.
However, despite its success in high-resource languages, instruction tuning remains a significant challenge for low-resource languages. One of the key bottlenecks is the scarcity of high-quality instruction-following datasets in these languages. Unlike English, where such resources are abundant, low-resource languages have very few. The process of manually creating instruction datasets is labor-intensive and expensive, often requiring native speakers with expertise in both the language and various task domains.  Consequently, researchers have frequently resorted to machine translation (MT) techniques to generate instruction data for these languages \citep{li2023bactrianxmultilingualreplicableinstructionfollowing, holmstrom-doostmohammadi-2023-making, li-etal-2024-x}. However, relying on MT to produce instruction tuning data introduces several complications. Translations may fail to capture nuanced meanings, cultural contexts, and idiomatic expressions inherent in the source language, leading to instruction-response pairs that are misaligned or unnatural in the target language \citep{bizzoni-etal-2020-human}. This misalignment can adversely affect the performance of LLMs trained on such data \citep{yu-etal-2022-translate}, as they may learn to generate responses that are semantically incorrect or culturally inappropriate.

Luxembourgish, a West Germanic language with about 400\,000 speakers in Luxembourg, exemplifies these challenges. As a low-resource language, it suffers from a paucity of linguistic data, making it difficult to develop robust LLMs or MT models.

To address the scarcity of high-quality data, we compile \textsc{LuxInstruct}, a cross-lingual instruction tuning dataset for Luxembourgish, in which instructions are written in other languages and outputs are in Luxembourgish. By avoiding machine translation into Luxembourgish\footnote{MT into Luxembourgish is limited to an ablation subset of instructions and omitted from \textsc{LuxInstruct}.}, our approach preserves linguistic integrity while enabling the adaptation of LLMs to Luxembourgish through alignment with English, French, and German. Additionally, the use of human-generated, rather than synthetic, data guarantees \textsc{LuxInstruct}'s high quality.

Our findings indicate that this cross-lingual dataset, due to its native construction, offers superior quality and has the potential to result in improved model performance compared to monolingual MT-based instruction tuning data.

\section{Related Work}
\subsection{Luxembourgish NLP}
Luxembourgish NLP is still in its early developmental phase. The field gained traction with the introduction of the encoder-only model \textsc{LuxemBERT} \citep{lothritz_luxembert_2022}, followed by the decoder-only \textsc{LuxGPT-2} \citep{bernardy2022luxembourgish}, and later the encoder-decoder models \textsc{LuxT5} and \textsc{LuxT5-Grande} \citep{plum-etal-2025-text}. Nonetheless, \citet{lothritz-etal-2026-testing} demonstrated that both open-source and many proprietary LLMs still fall short of achieving high-level performance in Luxembourgish.

In terms of existing datasets, the most substantial compilation of unlabeled Luxembourgish text to date was assembled by \citet{plum-etal-2025-text}, while \citet{philippy-etal-2025-luxembedder} contributed a parallel corpus covering English–Luxembourgish and French–Luxembourgish pairs. Nevertheless, a native high-quality instruction tuning dataset has yet to be developed.

\subsection{Low-Resource Language Instruction Tuning Data}
Although prior work has focused on creating instruction tuning datasets for specific languages \citep{masahiro2023japaneseinstructiondataset, azime-etal-2024-walia, laiyk2025instructiontuningpublicgovernment, shang-etal-2025-atlas}, many languages, including Luxembourgish, still lack such resources. 
This gap is largely due to the high cost of manually curating instruction tuning data for low-resource languages. Existing approaches typically rely on MT \citep{li2023bactrianxmultilingualreplicableinstructionfollowing, holmstrom-doostmohammadi-2023-making, li-etal-2024-x} or repurpose labeled NLP datasets \citep{muennighoff2023crosslingualgeneralizationmultitaskfinetuning}. However, neither method is effective for Luxembourgish, due to limited MT quality and a scarcity of labeled data.

\citet{koksal-etal-2024-longform} propose the use of \textit{reverse instructions} to generate instruction tuning data from raw text, a method later expanded to the \textit{Multilingual Reverse Instructions} (MURI) framework \citep{koksal-etal-2025-muri}. Yet, MURI still relies on two rounds of MT and focuses on multilingual (same-language) rather than cross-lingual (instruction and output in different languages) tuning. While multilingual tuning benefits low-resource settings \citep{weber-etal-2024-investigating, shaham-etal-2024-multilingual}, cross-lingual tuning has been shown to offer comparable advantages \citep{li-etal-2024-x, chai2025xcot, lin-etal-2025-crossin}.

\section{\textsc{LuxInstruct}}

\subsection{Dataset Creation}
We create LuxInstruct using three primary Luxembourgish data sources: Wikipedia, News Articles, and an Online Dictionary. More information on the source data and the process is provided in Appendix \ref{app:luxinstruct}.

\paragraph{Wikipedia}
We adopt a reverse instruction generation approach inspired by the MURI framework \citep{koksal-etal-2025-muri}, but diverge in key aspects to avoid translation artifacts. Instead of first translating the source data into English and then translating the generated instructions into the target language, as in MURI, we simply prompt OpenAI's \texttt{gpt-4.1-mini}\footnote{In preliminary tests, we found that \texttt{gpt-4.1-mini} cannot reliably generate Luxembourgish text, but reliably understands it well enough for this extraction task.} to extract informative content directly from Luxembourgish Wikipedia and generate corresponding instructions in English. This allows for high-quality semantically aligned instruction-output pairs without relying on MT. Additionally, unlike MURI, which applies a single prompt to full, often noisy documents, our method ensures cleaner inputs by allowing the model to select coherent spans. Generated pairs are further filtered based on a series of heuristic-based filtering steps (length, correct language, extraction consistency, etc.) to ensure data quality.
Then, we additionally machine-translate a subset of the instructions to German and French using \texttt{gpt-4.1-mini}. The resulting dataset forms the \texttt{\textbf{Open-Ended}} portion of \textsc{LuxInstruct}.

\paragraph{News Articles}
The Luxembourgish news platform \textit{RTL Luxembourg} publishes articles in Luxembourgish as well as French and English. Since there is no direct alignment between language versions, we use OpenAI's \texttt{text-embedding-3-small} model to retrieve bilingual article pairs (LB-EN \& LB-FR). From these article pairs, we create instruction–output pairs in two different task styles: (1) generating Luxembourgish news headlines from English or French articles (\texttt{\textbf{Article-To-Title}}), and (2) generating hypothetical Luxembourgish news articles from English or French headlines (\texttt{\textbf{Title-To-Article}}). In addition, similar monolingual Luxembourgish instruction-output pairs are created.

Furthermore, we leverage the \textsc{LuxAlign} \citep{philippy-etal-2025-luxembedder} corpus of parallel Luxembourgish-French-English semantically similar sentences derived from news articles to create a cross-lingual paraphrasing task (\texttt{\textbf{CL-Paraphrase}}).

\paragraph{Online Dictionary} 
We leverage a publicly available Luxembourgish dictionary containing lexical entries with synonyms, translations (to English, French, German) and example sentences. We design three task types: (1) generating Luxembourgish example sentences of a given Luxembourgish word, where the exact word meaning is given by the translation of the word (\texttt{\textbf{Word-To-Example}}), (2) simplifying colloquial Luxembourgish sentences (\texttt{\textbf{Colloquial-To-Standard}}), and (3) word translation to Luxembourgish (\texttt{\textbf{Word-Translation}}).

\subsection{Dataset Statistics}
Our new dataset consists of 391,551 cross-lingual instruction-output samples across English, French, and German as instruction languages, along with 145,793 monolingual samples in Luxembourgish, where both instruction and output are in Luxembourgish.
Appendix \ref{app:luxinstruct} provides the exact number of samples per language and type of task (Table \ref{tab:dataset_stats}) as well as examples (Table \ref{tab:examples}). 
\textsc{LuxInstruct} content derived from Wikipedia and the Online Dictionary is licensed under \textit{CC BY-SA 4.0}, whereas the News Articles content is non-commercial.

\section{Cross-Lingual Vs Monolingual Instruction Tuning} \label{sec:experiments}
We compare cross-lingual and monolingual instruction tuning for Luxembourgish through two experiments: one measuring cross-lingual alignment (\S \ref{sec:experiments_alignemnt}) and one evaluating instruction-following in in-context learning (\S \ref{sec:experiments_few_shot}). Both experiments use the \texttt{Open-Ended} portion, containing parallel English, French, and German instructions, further extended to Luxembourgish via \texttt{gpt-4.5}.

\subsection{Cross-Lingual Alignment} \label{sec:experiments_alignemnt}
Each model is fine-tuned on same-sized subsets, with instructions in a single language (EN, FR, DE, or LB) and responses in Luxembourgish.

We then assess the alignment between the Luxembourgish embedding space and the English, French, and German spaces by using parallel data from \textsc{Flores-200} \citep{nllbteam2022languageleftbehindscaling} and computing Centered Kernel Alignment (CKA) scores \citep{kornblith2019similarityneuralnetworkrepresentations} using the model's mean-pooled hidden states of its last layer. More technical details are provided in Appendix \ref{app:cross_lingual_alignment}.

Figure \ref{fig:alignment_results} shows the average increase in alignment between the Luxembourgish and the English, French, and German representation spaces after fine-tuning with different instruction languages. While alignment gains vary across models, cross-lingual instruction tuning proves at least as effective, and often more so, than monolingual tuning. EN-LB and FR-LB configurations yield the highest alignment improvements, whereas DE-LB performs often worse than monolingual (LB-LB) tuning. This suggests that pairing low-resource languages with more distant languages during instruction tuning may be more effective than using closely related ones. We provide the exact results per language pair in Table \ref{tab:full_alignment_results} in the Appendix.

\begin{figure}[h!]
    \centering
    \includegraphics[width=0.98\linewidth]{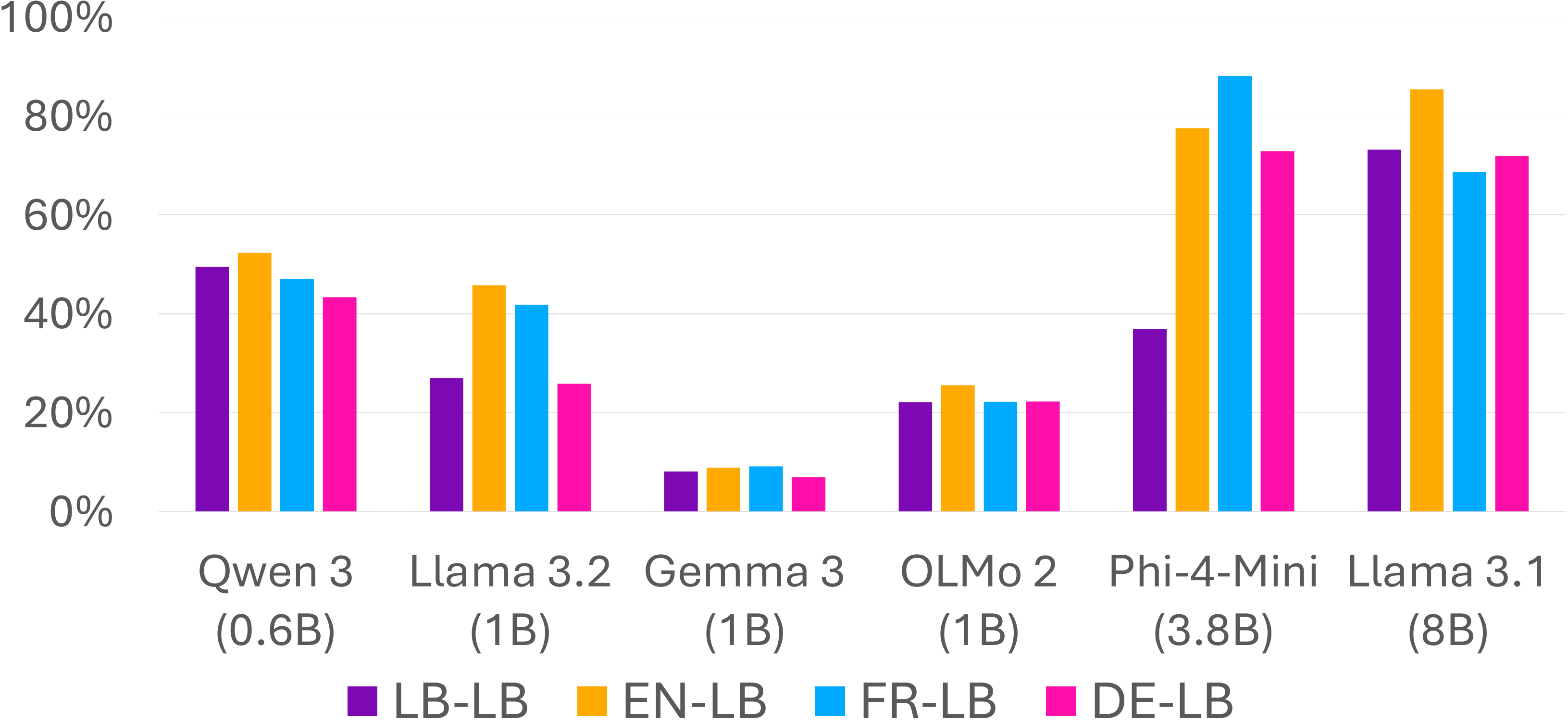}
    \caption{Mean variation (in \%) in alignment between the Luxembourgish and the English, French, and German representation spaces after fine-tuning on LB-LB, EN-LB, DE-LB, or FR-LB instruction tuning data}
    \label{fig:alignment_results}
\end{figure}

\subsection{Few-Shot In-Context Learning} \label{sec:experiments_few_shot}
To evaluate the impact of instruction language in instruction tuning on an LLM's text generation capabilities, we adopt a few-shot in-context learning strategy. More specifically, we compare different instruction languages (\texttt{LB}, \texttt{EN}, \texttt{FR}, \texttt{DE}, and a mixed-language setting \texttt{MULTI}) for these few-shot examples, while keeping the output language fixed to Luxembourgish.

We use \texttt{Gemma 3 12B} \& \texttt{27B} \citep{gemmateam2025gemma3technicalreport} for these experiments, as they are the only moderately sized open-source models among those tested that yield reasonable generation quality in Luxembourgish.

Given the absence of a standard instruction-following benchmark in Luxembourgish, we construct a set of 50 manually verified instructions. Since standard metrics are inapplicable, we employ the G-Eval framework \citep{liu-etal-2023-g}, with evaluations conducted via \texttt{GPT-5 mini}, \texttt{Gemini 2.5 Flash-Lite} \citep{comanici2025gemini25pushingfrontier}, and \texttt{DeepSeek-V3} \citep{deepseekai2025deepseekv3technicalreport}.
Using three LLM judges improves robustness, particularly in low-resource settings where evaluation may be less reliable than in English. To further validate this approach, Appendix \ref{app:human_evaluation} presents a human annotation study that verifies the consistency between human assessments and the aggregated judgments of the three LLM evaluators.

Table \ref{tab:few_shot_results_short} presents the average scores from all three evaluators across four different dimensions. Full model-specific scores are available in Table \ref{tab:few_shot_results_full} in the Appendix. To enhance result robustness, we repeat the experiments five times with different seeds (and corresponding few-shot examples), reporting the mean scores across all runs.

The findings clearly show that cross-lingual few-shot examples outperform monolingual Luxembourgish ones, demonstrating not only the utility of \textsc{LuxInstruct}, but also reinforcing our decision to pursue a cross-lingual strategy.

\begin{table}[h!]
\centering
\begin{tabular}{c|c|c|cccc}
\hline
 & \textbf{k} & \textbf{IT Lang.} & \textbf{CL} & \textbf{CO} & \textbf{FL} & \textbf{RE} \\
\hline

\multirow{11}{*}{\rotatebox[origin=c]{90}{Gemma 12B}} 
 & \multicolumn{2}{c|}{0-Shot} & 60.1 & 67.1 & 57.3 & 69.2 \\ \cline{2-7}
 & \multirow{5}{*}{4} & LB    & 61.6 & 70.9 & 59.4 & 74.0 \\
 &                         & DE    & 61.6 & 72.2 & 60.4 & 75.5 \\
 &                         & EN    & \textbf{63.4} & \textbf{73.2} & \textbf{61.8} & 76.0 \\
 &                         & FR    & 62.0 & 72.6 & 60.7 & \textbf{76.1} \\
 &                         & MULTI & 61.6 & 71.8 & 59.4 & 74.5 \\ \cline{2-7}
 & \multirow{5}{*}{8} & LB    & 62.1 & 72.1 & 60.3 & 75.7 \\
 &                         & DE    & 61.7 & 72.7 & 60.8 & 76.3 \\
 &                         & EN    & \textbf{64.7} & \textbf{74.8} & \textbf{63.0} & \textbf{77.9} \\
 &                         & FR    & 62.6 & 73.2 & 61.9 & 76.9 \\
 &                         & MULTI & 62.4 & 72.9 & 61.2 & 76.5 \\
\hline

\multirow{11}{*}{\rotatebox[origin=c]{90}{Gemma 27B}} 
 & \multicolumn{2}{c|}{0} & 68.9 & 74.2 & 67.1 & 75.1 \\ \cline{2-7}
 & \multirow{5}{*}{4} & LB    & 77.6 & 83.5 & 75.6 & 85.2 \\
 &                         & DE    & 76.6 & 83.0 & 74.7 & 85.0 \\
 &                         & EN    & \textbf{78.7} & \textbf{85.1} & \textbf{77.5} & \textbf{86.8} \\
 &                         & FR    & 77.1 & 84.3 & 74.8 & 85.3 \\
 &                         & MULTI & 78.1 & 84.0 & 75.7 & 85.5 \\ \cline{2-7}
 & \multirow{5}{*}{8} & LB    & 78.9 & 85.8 & 76.4 & 87.1 \\
 &                         & DE    & 80.5 & 87.8 & 78.4 & \textbf{89.6} \\
 &                         & EN    & \textbf{81.3} & \textbf{88.7} & \textbf{79.9} & 89.3 \\
 &                         & FR    & 78.1 & 85.6 & 76.8 & 87.3 \\
 &                         & MULTI & 80.1 & 86.4 & 78.0 & 88.2 \\
\hline
\end{tabular}

\caption{G-Eval results measuring instruction-following performance in Luxembourgish across Clarity (\textbf{CL}), Coherence (\textbf{CO}), Fluency (\textbf{FL}), and Relevance (\textbf{RE}) in 0-shot, 4-shot, and 8-shot settings, with few-shot examples using different instruction languages (\textbf{IT Lang.}).}
\label{tab:few_shot_results_short}
\end{table}

% \begin{table}[h!]
% \centering
% \begin{tabular}{c|c|cccc}
% \hline
%  & IT Lang. & \textbf{CL} & \textbf{CO} & \textbf{FL} & \textbf{RE} \\
% \hline
% \multicolumn{2}{c|}{0-Shot} & 67.27 & 72.92 & 65.41 & 74.81 \\ \hline
% \multirow{5}{*}{\rotatebox[origin=c]{90}{4-Shot}} & LB & 68.50 & 76.92 & 66.71 & 80.18 \\
%  & DE    & 69.12 & 78.06 & 67.84 & 80.94 \\
%  & EN    & \textbf{71.53} & \textbf{79.27} & \textbf{70.48} & \textbf{82.03} \\
%  & FR    & 69.73 & 78.84 & 68.71 & 81.95 \\
%  & MULTI & 68.91 & 77.87 & 67.23 & 80.80 \\
% \hline
% \multirow{5}{*}{\rotatebox[origin=c]{90}{8-Shot}} & LB & 69.24 & 78.37 & 67.94 & 81.29 \\
%  & DE    & 69.26 & 78.81 & 68.54 & 82.14 \\
%  & EN    & \textbf{72.19} & \textbf{80.38} & \textbf{71.61} & \textbf{83.30} \\
%  & FR    & 69.92 & 79.08 & 69.73 & 82.42 \\
%  & MULTI & 69.59 & 78.51 & 68.72 & 81.91 \\
% \hline
% \end{tabular}
% \caption{G-Eval results measuring instruction-following performance in Luxembourgish across Clarity (\textbf{CL}), Coherence (\textbf{CO}), Fluency (\textbf{FL}), and Relevance (\textbf{RE}) in 0-shot, 4-shot, and 8-shot settings, with few-shot examples using different instruction languages (\textbf{IT Lang.}).}
% \label{tab:few_shot_results_short}
% \end{table}
\section{Discussion}
We believe that the construction of \textsc{LuxInstruct} represents a significant step forward for resource development in Luxembourgish. Its human-written outputs ensure natural, reliable targets, while the cross-lingual design aids alignment across languages (Section \ref{sec:experiments}).

Its most immediate application is improving the linguistic accuracy and fluency of models in Luxembourgish, particularly in grammar, orthography, and stylistic consistency. At the same time, the dataset plays a broader role by embedding a culturally grounded and context-aware understanding of the language within LLMs.

To this end, Wikipedia serves as a curated source of both global and local knowledge. The Luxembourgish edition emphasizes nationally relevant topics, including local history, government institutions, prominent cultural figures, and regional traditions.

This foundation is further strengthened by news articles, which provide insight into the current sociopolitical and cultural landscape of Luxembourg. News sources reflect real-time discourse and events, anchoring language use in a dynamic, evolving context. This allows models to produce outputs that are timely, accurate, and contextually informed.

Additionally, lexical and dictionary resources introduce an essential semantic layer. The dictionaries used include multilingual translations and clarifying examples for polysemous terms, helping the model capture context-specific meanings and reduce ambiguity.

While the current dataset forms a foundation for instruction tuning in Luxembourgish, future efforts will focus on scaling this work. We aim to apply \textsc{LuxInstruct} on a larger scale to existing LLMs, further enriching their Luxembourgish capabilities. Parallel to this, we will continue expanding the dataset in both size and diversity, incorporating new seed sources and adopting emerging data generation techniques.

\section{Conclusion}
This work presents the development of \textsc{LuxInstruct}, the first cross-lingual instruction tuning dataset tailored for Luxembourgish. By incorporating instructions in English, French and German, the dataset enables cross-lingual model alignment for Luxembourgish. Moreover, we provide empirical evidence demonstrating the advantages of cross-lingual instruction tuning over monolingual approaches in such settings. We hope that both the dataset and our findings serve as a foundation for further advancements in Luxembourgish NLP.
\section*{Limitations}
The key limitation of our dataset is its restricted diversity, as it currently covers only seven task types. This constraint reflects the scarcity of high-quality Luxembourgish resources. We made a conscious decision to prioritize quality—relying on human-generated seed data—over quantity or breadth, avoiding large-scale translation from high-resource languages which often introduces noise or mistranslations.

To add some variation, we included multilingual instructions (in three languages plus Luxembourgish) and varied instruction templates where possible. We view this dataset as a starting point and plan to expand it in future work by incorporating additional tasks and further increasing linguistic and instructional diversity.
\section*{Ethical Considerations}
Our dataset is constructed from publicly available sources, including news articles and Wikipedia entries, which may contain the names of individuals. We chose not to anonymize this information, as doing so would significantly reduce the contextual richness of the data. Since the content is already accessible in the public domain, we consider its inclusion ethically permissible. Preserving these references is important for maintaining data integrity and ensuring the effectiveness of the dataset for real-world applications.

%\section*{Acknowledgments}

\bibliography{latex/anthology-1, latex/custom}

\appendix

\section{\textsc{LuxInstruct}} \label{app:luxinstruct}
\subsection{Creation Process}
Here we provide further details on how \textsc{LuxInstruct} was constructed using 3 different sources: 1) Wikipedia, 2) News Articles, 3) Online Dictionary. These sources contain high-quality Luxembourgish text and are inherently safety-filtered, removing the need for an additional safety filtering step.

\subsubsection{Wikipedia}
For the \texttt{Open-Ended} component of \textsc{LuxInstruct}, we use the Luxembourgish subset of the Wikipedia dumps \citep{wikidump}\footnote{\url{https://huggingface.co/datasets/wikimedia/wikipedia/viewer/20231101.lb}}, released under the \textit{CC BY-SA 4.0} license\footnote{\url{https://creativecommons.org/licenses/by-sa/4.0/}}. Out of roughly 64,000 articles, we randomly select about 20,000 to generate our samples.

The English instruction generation is carried out using \texttt{gpt-4.1-mini}, which was selected due to its computational efficiency and cost-effectiveness. In a small manual qualitative preliminary evaluation, we found that it demonstrates sufficiently strong Luxembourgish comprehension. As the task requires only understanding, not generating, Luxembourgish text, its generative capabilities in Luxembourgish are not a priority. To ensure reliable extraction of the generated pairs and to avoid formatting inconsistencies, we use function calling with a predefined JSON schema to structure the model’s responses in a machine-readable format. The model is guided by the following prompt:

\begin{tcolorbox}[
    colback=gray!20, % Background color (20% grey)
    colframe=gray!80, % Border color
    boxrule=0.5mm, % Border thickness
    arc=4mm, % Rounded corners
    auto outer arc, % Apply rounded corners to outer frame
    width=\linewidth, % Width of the box
    enlarge left by=0mm, % Optional: adjust box margin
    enlarge right by=0mm, % Optional: adjust box margin
    ]

{\ttfamily Create structured instruction- tuning data for language models. From the text below, extract coherent excerpts that a model might generate in response to a clear, concise instruction. Each excerpt should be a complete, accurate, and natural language response and should be taken directly from the text without altering it. Instructions must be in English, answers in Luxembourgish. Ensure instructions are self-contained and context-independent. The instruction does not need to specify Luxembourgish as the output language. \\

Input Text: \\

\{text\} \\

Return your findings in JSON format.}
\end{tcolorbox}

After generating the English instructions, we apply a series of simple heuristic-based filters to remove low-quality instruction-output pairs. A sample is discarded if it meets any of the following criteria:

\begin{itemize}
\setlength\itemsep{0.1em}
    \item The output is not a string;
    \item The output contains fewer than 10 words;
    \item The instruction contains the word \texttt{List};
    \item The output begins with a lowercase letter;
    \item The output contains a question mark;
    \item The output does not end with a full stop;
    \item The output is not written in Luxembourgish;
    \item The output is not present in the original Wikipedia article (determined via fuzzy matching, allowing for minor inconsistencies such as punctuation differences or sentence truncation).
\end{itemize}

While we use a language identification model to ensure outputs are predominantly Luxembourgish, natural code-switching is intentionally preserved, as it is inherent to authentic Luxembourgish usage.

\subsubsection{News Articles}
We collect articles from RTL\footnote{\url{https://www.rtl.lu}}, a Luxembourgish news platform publishing in Luxembourgish, as well as in French since 2011 and in English since 2018. Since each language has a separate website, articles are not explicitly aligned across languages. To find matching articles, we encode them using OpenAI’s \texttt{text-embedding-3-small} and select pairs with cosine similarity above 0.65.
We discard articles that are too short or too long, and those with titles under six words. The resulting article pairs are used to build the \texttt{\textbf{Article-To-Title}} and \texttt{\textbf{Title-To-Article}} datasets in \textsc{LuxInstruct}.

We also create 50 prompt templates in each language (e.g., "\texttt{Draft a publication-ready article based on the headline provided:}") and randomly assign one to each pair.

We apply a similar procedure, excluding the cross-lingual article matching, to Luxembourgish-only articles in order to construct the monolingual portions of \texttt{\textbf{Article-To-Title}} and \texttt{\textbf{Title-To-Article}}.

The \texttt{\textbf{CL-Paraphrase}} component leverages the \textsc{LuxAlign} dataset \citep{philippy-etal-2025-luxembedder}, which includes 28.2k Luxembourgish-English and 89.4k Luxembourgish-French semantically aligned sentence pairs. Here too, we generate 50 prompt templates (e.g., "\texttt{Modify the sentence with new wording:}") and assign them randomly to each bilingual pair, where the input is either English or French and the output is Luxembourgish.

\subsubsection{Online Dictionary}
The \textit{Luxembourg Online Dictionary} (LOD) provides free online access to Luxembourgish vocabulary, including translations into four languages—French, German, English, and Portuguese—as well as contextual usage examples for numerous terms. The dataset is fully accessible online\footnote{\url{https://data.public.lu/en/datasets/letzebuerger-online-dictionnaire-lod-linguistesch-daten/}} under the \textit{CC0 1.0} license\footnote{\url{https://creativecommons.org/publicdomain/zero/1.0/}}.

All \textsc{LuxInstruct} components sourced from LOD, namely \texttt{\textbf{Colloquial-To-Standard}}, \texttt\textbf{{Word-To-Example}}, and \texttt{\textbf{Word-Translation}}, are generated using 20 prompt templates per language for the first task and 50 prompt templates per language for the latter two tasks.

\subsection{Dataset Statistics}
\begin{table*}[h!]
\centering
\begin{tabular}{lrrrr|r}
\hline
\textbf{} & \textbf{de} & \textbf{fr} & \textbf{en} & \textbf{lb} & \textbf{Total} \\
\hline
\texttt{\textbf{Article-To-Title}}    &        & 11\,774  & 4\,744   & 51\,498  & 68\,016 \\
\texttt{\textbf{Title-To-Article}} & & 12\,891  & 4\,899   & 51\,507  &  69\,297 \\
\texttt{\textbf{Colloquial-To-Standard}}    & 2\,323   & 2\,323   & 2\,323   & 2\,323   & 9\,292 \\
\texttt{\textbf{Word-To-Example}}         & 40\,426  & 40\,390  & 38\,418  & 40\,465  & 159\,699 \\
\texttt{\textbf{Open-Ended}}            & 19\,955  & 19\,955  & 56\,633  &   & 96\,543 \\
\texttt{\textbf{Word-Translation}}            & 6\,153  & 5\,840  & 4\,927  &  & 16\,920 \\
\texttt{\textbf{CL-Paraphrase}} &   & 89\,405  & 28\,172 &  & 117\,577 \\
\hline
\textbf{Total} & 68\,857  & 182\,578  & 140\,116 & 145\,793 & 537\,344 \\
\hline
\end{tabular}
\caption{Data distribution across different languages and task types}
\label{tab:dataset_stats}
\end{table*}
The exact numbers of created samples per instruction language and per task type are provided in Table \ref{tab:dataset_stats}.

\subsection{Examples from \textsc{LuxInstruct}} \label{app:examples}

Table \ref{tab:examples} contains 2 examples per task type.

\section{Experimental Setup}
We conduct our experiments on a subset of the \texttt{Open-Ended} section of \textsc{LuxInstruct}, restricted to the samples where instructions have additionally been translated into French and German. For the purposes of these experiments, we additionally translate these instructions into Luxembourgish using \texttt{gpt-4.5}, though these translations are not included in the final dataset.

\subsection{Cross-Lingual Alignment}
\label{app:cross_lingual_alignment}

\subsubsection{Models}
In our experiments we use the following models: \\

\noindent \textbf{Qwen3-0.6B}\footnote{\url{https://huggingface.co/Qwen/Qwen3-0.6B}} \citep{yang2025qwen3technicalreport} \\
A 0.75B‑parameter, 28-layer, instruction‑tuned model with a 32K context window, trained with 36T tokens and with multilingual support in over 119 languages, including Luxembourgish, released under the \textit{Apache 2.0} license\footnote{\url{https://www.apache.org/licenses/LICENSE-2.0}}.

\noindent \textbf{Gemma-3-1B-IT}\footnote{\url{https://huggingface.co/google/gemma-3-1b-it}} \citep{gemmateam2025gemma3technicalreport} \\
A 1B‑parameter, 26-layer, instruction‑tuned model with a 32K context window, trained with 2T tokens and with multilingual support in over 140 languages, including Luxembourgish, released with the \textit{Gemma Terms of Use}\footnote{\url{https://ai.google.dev/gemma/terms}}.

\noindent \textbf{OLMo-2-1B-Instruct}\footnote{\url{https://huggingface.co/allenai/OLMo-2-0425-1B-Instruct}} \citep{olmo20252olmo2furious} \\
A 1.48B‑parameter, 16-layer, instruction‑tuned model, trained with 4T tokens, with primarily English support, released under the \textit{Apache 2.0} license\footnote{\url{https://www.apache.org/licenses/LICENSE-2.0}}.

\noindent \textbf{Llama-3.2-1B-Instruct}\footnote{\url{https://huggingface.co/meta-llama/Llama-3.2-1B-Instruct}} \\
A 1.23B‑parameter, 16-layer, instruction‑tuned model with a 128K context window, trained with 5T tokens and with multilingual support in 8 languages, released under the \textit{Llama 3.2 Community License}\footnote{\url{https://www.llama.com/llama3_2/license/}}.

\noindent \textbf{Phi-4-Mini-Instruct}\footnote{\url{https://huggingface.co/microsoft/Phi-4-mini-instruct}} \citep{microsoft2025phi4minitechnicalreportcompact} \\
A 3.84B‑parameter, 32-layer, instruction‑tuned model with a 128K context window, trained with 9T tokens and with multilingual support in 23 languages, released under the \textit{MIT License}\footnote{\url{https://opensource.org/license/mit}}.

\noindent \textbf{Llama-3.1-8B-Instruct}\footnote{\url{https://huggingface.co/meta-llama/Llama-3.1-8B-Instruct}} \citep{grattafiori2024llama3herdmodels} \\
A 8.03B‑parameter, 32-layer, instruction‑tuned model with a 128K context window, trained with 9T tokens and with multilingual support in 8 languages, released under the \textit{MIT License}\footnote{\url{https://www.llama.com/llama3_1/license/}}.

\subsubsection{Technical Details}
We apply LoRA \citep{hu2021loralowrankadaptationlarge} to fine-tune the value, query, and key projections in the attention layers, using a rank of 8, scaling factor $\alpha = 16$, and a dropout rate of 0.05. Each model is trained for 500 steps with a batch size of 16, a learning rate of 2e$^{-5}$, weight decay of 0.01, and a context length of 128 tokens. 

To compute cross-lingual alignment scores between language pairs, we use the \textit{devtest} split of the \textsc{Flores-200} dataset\footnote{\url{https://huggingface.co/datasets/facebook/flores}} \citep{nllbteam2022languageleftbehindscaling}, which contains 1\,012 parallel sentences across 204 languages, including Luxembourgish. Document-level representations are obtained by mean-pooling the final-layer contextualized token embeddings. Alignment between languages is quantified using the Centered Kernel Alignment (CKA) metric \citep{kornblith2019similarityneuralnetworkrepresentations}.

All experiments are conducted on a single Nvidia T4 GPU and complete within a few hours.

\subsubsection{Full Results}
In Table \ref{tab:full_alignment_results} we provide the full results that have been summarized in Figure \ref{fig:alignment_results}.

\subsection{Few-Shot In-Context Learning}
\label{app:few_shot_in_context}

\subsubsection{Text Generation Model}
In this experiment we use \textbf{\texttt{Gemma 3}} \citep{gemmateam2025gemma3technicalreport} in its 12B and 27B versions, with 48 and 62 layers and trained on 12T and 14T tokens, respectively. Both models are instruction-tuned with a 128K context window, support over 140 languages (including Luxembourgish), and include visual capabilities.

\subsubsection{Test Data}
Due to the absence of high-quality instruction-following benchmarks in Luxembourgish, a native speaker was engaged to curate a set of 50 Luxembourgish test instructions, which were subsequently used to evaluate the model's performance.\footnote{The exact test instructions are provided \href{https://github.com/fredxlpy/LuxInstruct}{here}.}

\subsubsection{Technical Details}
To accommodate resource limitations, the model used for text generation is quantized to 4-bit precision. We perform generation using parameters $top_k=64$, $top_p=0.95$, and temperature $=1.0$, with a maximum output length of 500 tokens.

Since we machine-translated a subset of the instructions from the \texttt{Open-Ended} section of \textsc{LuxInstruct}, we are able to use the same few-shot in-context examples across all instruction language settings.

This design choice enables us to isolate the influence of instruction language while controlling for few-shot examples selection variability. To further enhance the reliability of our results, we perform five independent generations using five distinct sets of few-shot examples, and incorporate all generated outputs in the subsequent evaluation.

In the \texttt{LB}, \texttt{FR}, \texttt{EN}, and \texttt{DE} settings, all few-shot instructions are provided solely in the respective languages, with outputs always in Luxembourgish. In contrast, the \texttt{MULTI} setting combines instructions across languages: one per language in the 4-shot setup and two per language in the 8-shot setup.

\subsubsection{G-Eval}

For evaluation, we adopt the G-Eval framework \citep{liu-etal-2023-g}, which employs an LLM-as-a-judge approach. We use its implementation via the \texttt{DeepEval} library \citep{deepeval2025}, using it to assess outputs along four dimensions: Clarity, Relevance, Fluency, and Coherence. Scores are returned on a 0–1 scale, which we multiply by 100 for reporting. To enhance robustness, we conduct evaluations using three different LLM APIs: \texttt{GPT-5 mini}, \texttt{Gemini 2.5 Flash-Lite} \citep{comanici2025gemini25pushingfrontier}, and \texttt{DeepSeek-V3} \citep{deepseekai2025deepseekv3technicalreport}.

The specific Chain-of-Thought prompts used across the four evaluation dimensions are as follows:

\begin{tcolorbox}[mygraybox, title=Clarity]
{\ttfamily \small
\begin{itemize}[left=0pt, labelsep=5pt, itemsep=2pt, parsep=0pt]
    \item Compare the language used in the output with the input to assess clarity and simplicity.
    \item Check if the output presents information logically and in an organized manner relative to the input.
    \item Identify any ambiguous or unclear phrases in the output that deviate from the meaning of the input.
    \item Ensure that the formatting and structure of the output enhance understanding compared to the input.
\end{itemize}
}
\end{tcolorbox}

\begin{tcolorbox}[mygraybox, title=Relevance]
{\ttfamily \small
\begin{itemize}[left=0pt, labelsep=5pt, itemsep=2pt, parsep=0pt]
    \item Compare the output to the input instruction to determine if the output addresses the task requested.
    \item Verify that the content of the output is contextually relevant and directly related to the input instruction.
    \item Check for completeness by ensuring the output fully responds to the elements specified in the input instruction.
    \item Assess clarity and coherence of the output in relation to the input instruction, confirming it makes sense as a response.
\end{itemize}
}
\end{tcolorbox}

\begin{tcolorbox}[mygraybox, title=Fluency]
{\ttfamily \small
\begin{itemize}[left=0pt, labelsep=5pt, itemsep=2pt, parsep=0pt]
    \item Compare the output to the input to ensure the content is relevant and appropriately reflects the input context.
    \item Check the grammar in the output for correctness including verb tense, subject-verb agreement, and sentence structure.
    \item Evaluate the coherence of the output by assessing logical flow and clarity throughout the response.
    \item Determine if the output reads naturally and smoothly as a fluent piece of text in relation to the input.
\end{itemize}
}
\end{tcolorbox}

\begin{tcolorbox}[mygraybox, title=Coherence]
{\ttfamily \small
\begin{itemize}[left=0pt, labelsep=5pt, itemsep=2pt, parsep=0pt]
    \item Compare the input with the output to determine if the output logically addresses the content and intent of the input.
    \item Assess the structure of the output for clear organization and progression of ideas relative to the input.
    \item Evaluate whether the output maintains a consistent and coherent flow that aligns with the input's context and purpose.
    \item Identify any gaps or contradictions between the input and output that disrupt logical coherence.
\end{itemize}
}
\end{tcolorbox}

We provide the full evaluation results in Table \ref{tab:few_shot_results_full}.

\subsubsection{Human Calibration of G-Eval} \label{app:human_evaluation}
% Motivated by the concern that relying on a single judge LLM may be risky in low-resource settings such as Luxembourgish, we employed three independent judge LLMs. 

To assess the reliability of LLM-based judgments, we complemented our automatic evaluation with a small-scale human study. Our working assumption is that the consensus formed when all three judge LLMs prefer the same outputs offers a more stable and trustworthy indicator of quality than using only one model. To evaluate this assumption, we randomly sampled output pairs on which all judge LLMs agreed about the preferred candidate (i.e., one output consistently received higher scores across dimensions). Human annotators were then asked to assess the same pairs, without any knowledge of the LLMs' choice. To reduce annotation effort, we merged highly correlated evaluation dimensions: clarity and fluency were combined into one metric, while relevance and coherence formed another. For each pair, the annotator chose between three options: preference for the first output, preference for the second, or a tie. We then measured agreement between the human annotations and the LLM judges using Cohen's $\kappa$. Since the LLMs do only rarely produce ties (due to the continuous nature of their scores), tie cases from the human annotations were excluded from the $\kappa$ calculation.

\begin{table}[h]
    \centering
    \renewcommand{\arraystretch}{1.5}
    \begin{tabular}{c|c|c}
        \parbox{2cm}{\centering \textbf{Human} \\ \textbf{Annotator}} & \textbf{CL \& FL} & \textbf{RE \& CO} \\[ 5pt] \hline
        A & 0.957 & 0.935 \\
        B & 0.863 & 0.780 \\ \hline
    \end{tabular}
    \caption{Cohen’s $\kappa$ scores measuring agreement between two human annotators and the LLM judges for two sets of dimensions: Clarity and Fluency (CL \& FL), and Relevance and Coherence (RE \& CO)}
    \label{tab:human_evaluation}
\end{table}

\begin{table*}[ht]
\centering
\renewcommand{\arraystretch}{2}
\small
\begin{tabular}{c|p{6cm}|p{6cm}|}
\hline
\multirow{2}{*}{\textbf{Task Type}} & \multicolumn{2}{c|}{\textbf{Example}} \\ \cline{2-3} 
 & \multicolumn{1}{c|}{\textbf{Instruction}} & \multicolumn{1}{c|}{\textbf{Output}} \\ \hline
\multirow{2}{*}{\texttt{\textbf{Open-Ended}}} & Since when has Murcia been an autonomous community? & Murcia ass zanter dem 9. Juni 1982 eng autonom Gemeinschaft. \\ \cdashline{2-3}
 & Who was Thomas Keith Glennan and what was his role at NASA? & Den Thomas Keith Glennan, gebuer den 8. September 1905 zu Enderlin am North Dakota, USA, a gestuerwen den 11. Abrëll 1995 zu Mitchellville, Maryland, war tëscht August 1958 a Januar 1961 den éischten Administrateur vun der US-Raumfaartagentur NASA. \\ \hline
\multirow{2}{*}{\texttt{\textbf{Article-To-Title}}} & Turn the following news article into a concise headline: "Greece will have to be patient before it gets the next installment of its European financial aid. Following a meeting of the EU ministers..." & Griicheland muss sech nach e bësse gedëllegen \\ \cdashline{2-3}
 & Pick a headline that would summarize the article below: "Several commemoration ceremonies took place last Sunday, in order to celebrate Luxembourg’s national resistance day. The commemoration ceremony doesn’t vary much over the years. However, recent images..." & Erënnerungen u Krich an Ënnerdréckung héichhalen \\ \hline
\multirow{2}{*}{\texttt{\textbf{Title-To-Article}}} & Use this news headline to inspire a detailed article: "Social media to comply with new EU regulations" & Video-Plattforme wéi Youtube musse sech an der EU an Zukunft u méi strikt Reegele beim Jugendschutz oder och bei Reklammen halen. Déi zoustänneg Kommissioun vum Europaparlament... \\ \cdashline{2-3}
 & Produce an informative and factual story using this title: "Police looking for driver involved in pedestrian hit and run" & Zu Wolz gouf eng Persoun op engem Zebrasträife ugestouss. Ouni sech ëm d'Affer ze këmmeren, ass den Auto einfach fortgefuer. En Donneschdeg de Moien um kuerz virun 11 Auer... \\\hline
\multirow{2}{*}{\texttt{\textbf{Word-To-Example}}} & Demonstrate usage of the Luxembourgish word "eethesch" (translation: "ethical") in a sentence. & Den Asaz vu Kënschtlecher Intelligenz bréngt dacks eethesch a sozial Erausfuerderunge mat sech. \\ \cdashline{2-3}
 & Use the term "Fuerscherin" in a Luxembourgish sentence, translating to "researcher". & Déi jonk Fuerscherin sicht mat hirem Team no Léisunge géint de Klimawandel. \\\hline
\multirow{2}{*}{\parbox[c]{2.5cm}{\centering \texttt{\textbf{Colloquial-}}\\\texttt{\textbf{To-}}\\\texttt{\textbf{Standard}}}} & Clarify the meaning of this informal sentence: "Him ass eng gutt Geleeënheet laanscht d'Nues gaangen." & Hien huet eng gutt Geleeënheet verpasst. \\ \cdashline{2-3}
 & Rephrase the following colloquial sentence to make it easier to understand: "Den Informatiker huet de Computer mat Date gefiddert." & Den Informatiker huet Daten an de Computer aginn. \\ \hline
 \multirow{2}{*}{\texttt{\textbf{CL-Paraphrase}}} & Rework this sentence in a new form: "Luxembourg's population increased by 10,667 people in 2021." & D'Awunnerzuel zu Lëtzebuerg ass 2021 ëm 10.667 Persounen eropgaangen. \\ \cdashline{2-3}
 & Paraphrase the following: "He had been in shock himself, under a lot of stress and scared." & Hie selwer hätt ënner Schock, Stress an Angscht gestanen. \\ \hline
  \multirow{2}{*}{\texttt{\textbf{Word-Translation}}} & Find 2 Luxembourgish translations for the English word "appreciation [judgement, evaluation]". & "Aschätzung" an "Appreciatioun" \\ \cdashline{2-3}
 & List 3 ways "consent" can be said in Luxembourgish. & "Autorisatioun", "Geneemegung" an "Awëllegung" \\ \hline
\end{tabular}
\caption{Examples from \textsc{LuxInstruct} for each task type}
\label{tab:examples}
\end{table*}

\begin{table*}[ht]
\renewcommand{\arraystretch}{1.3}
\centering
\begin{tabular}{|c|c|c|c|c|}
\hline
\multirow{2}{*}{\textbf{Model}} & \multirow{2}{*}{\textbf{Training Data}} & \multicolumn{3}{c|}{\textbf{Compared embedding spaces}} \\
\cline{3-5}
 &  & \textbf{LB-DE} & \textbf{LB-EN} & \textbf{LB-FR} \\
\hline
\multirow{5}{*}{Qwen 3 (0.6B)} 
 & Base    & 0.2612 & 0.2342 & 0.2198 \\
 & DE-LB   & 0.3542 & 0.3456 & 0.3257 \\
 & EN-LB   & \textbf{0.3909} & \textbf{0.3610} & \textbf{0.3378} \\
 & FR-LB   & 0.3894 & 0.3587 & 0.3033 \\
 & LB-LB   & 0.3822 & 0.3528 & 0.3347 \\
\hline
\multirow{5}{*}{Llama 3.2 (1B)} 
 & Base    & 0.2359 & 0.2155 & 0.2091 \\
 & DE-LB   & 0.2649 & 0.2912 & 0.2754 \\
 & EN-LB   & \textbf{0.3332} & \textbf{0.3222} & \textbf{0.3076} \\
 & FR-LB   & 0.3302 & 0.3197 & 0.2871 \\
 & LB-LB   & 0.2950 & 0.2759 & 0.2678 \\
\hline
\multirow{5}{*}{Gemma 3 (1B)} 
 & Base    & 0.2774 & 0.2303 & 0.2144 \\
 & DE-LB   & 0.2981 & 0.2488 & 0.2255 \\
 & EN-LB   & 0.3029 & \textbf{0.2570} & 0.2264 \\
 & FR-LB   & \textbf{0.3035} & 0.2555 & 0.2290 \\
 & LB-LB   & 0.3027 & 0.2490 & \textbf{0.2292} \\
\hline
\multirow{5}{*}{OLMo 2 (1B)} 
 & Base    & 0.3020 & 0.2666 & 0.2784 \\
 & DE-LB   & 0.3381 & 0.3544 & 0.3429 \\
 & EN-LB   & 0.3639 & 0.3555 & \textbf{0.3438} \\
 & FR-LB   & \textbf{0.3682} & \textbf{0.3665} & 0.3003 \\
 & LB-LB   & 0.3582 & 0.3384 & 0.3376 \\
\hline
\multirow{5}{*}{Phi-4-Mini (1.8B)} 
 & Base    & 0.2090 & 0.1787 & 0.1879 \\
 & DE-LB   & 0.3494 & 0.3262 & 0.3196 \\
 & EN-LB   & 0.3600 & 0.3345 & 0.3273 \\
 & FR-LB   & \textbf{0.3815} & \textbf{0.3547} & \textbf{0.3466} \\
 & LB-LB   & 0.2798 & 0.2541 & 0.2539 \\
\hline
\multirow{5}{*}{Llama 3.1 (8B)} 
 & Base    & 0.2620 & 0.2278 & 0.2381 \\
 & DE-LB   & 0.4046 & 0.4177 & 0.4292 \\
 & EN-LB   & \textbf{0.4747} & \textbf{0.4315} & \textbf{0.4433} \\
 & FR-LB   & 0.4496 & 0.4075 & 0.3709 \\
 & LB-LB   & 0.4520 & 0.3975 & 0.4113 \\
\hline
\end{tabular}
\caption{CKA values for various models and training data configurations. Bold values indicate the highest per column within each model.}
\label{tab:full_alignment_results}
\end{table*}

\begin{table*}
\centering
\renewcommand{\arraystretch}{1.2}
\begin{tabular}{c|c|c|cccc|cccc}
\hline
\multirow{2}{*}{\textbf{Judge}} & \multirow{2}{*}{\textbf{Shots}} & \multirow{2}{*}{\textbf{IT Lang.}} &  \multicolumn{4}{c|}{\textbf{Gemma 3 12B}} & \multicolumn{4}{c}{\textbf{Gemma 3 27B}} \\
 &  &  & CL & CO & FL & RE & CL & CO & FL & RE \\
\hline

\multirow{11}{*}{\rotatebox[origin=c]{90}{DeepSeek-V3}} 
 & \multicolumn{2}{c|}{0-Shot} & 78.04 & 80.20 & 77.64 & 81.96 & 76.64 & 78.56 & 79.40 & 76.56 \\ \cline{2-11}
 & \multirow{5}{*}{4-Shot} 
   & LB    & 81.16 & 85.36 & 81.00 & 87.40 & 88.36 & 89.40 & 91.48 & 90.52 \\
 & & DE    & 81.84 & 86.92 & 82.28 & \textbf{89.72} & 88.24 & 87.96 & 91.32 & 90.68 \\
 & & EN    & 82.32 & 86.88 & \textbf{82.92} & 89.20 & \textbf{88.92} & 90.40 & \textbf{92.76} & \textbf{91.56} \\
 & & FR    & \textbf{82.36} & \textbf{86.96} & 82.88 & 89.44 & 88.44 & \textbf{91.28} & 91.16 & 90.12 \\
 & & MULTI & 82.04 & 85.96 & 81.84 & 88.68 & 88.28 & 89.40 & 90.88 & 89.92 \\ \cline{2-11}
 & \multirow{5}{*}{8-Shot} 
   & LB    & 82.32 & 86.88 & 82.28 & 89.40 & 91.16 & 92.16 & 92.84 & 93.96 \\
 & & DE    & 81.96 & 86.68 & 83.16 & 90.20 & 92.80 & 94.68 & 95.24 & \textbf{96.16} \\
 & & EN    & \textbf{83.64} & \textbf{88.40} & \textbf{84.64} & \textbf{91.00} & \textbf{93.76} & \textbf{94.88} & \textbf{95.52} & 95.40 \\
 & & FR    & 82.24 & 86.88 & 83.64 & 90.32 & 90.56 & 92.48 & 93.88 & 93.92 \\
 & & MULTI & 83.04 & 87.32 & 83.68 & 90.36 & 91.68 & 92.04 & 93.44 & 93.96 \\
\hline

\multirow{11}{*}{\rotatebox[origin=c]{90}{Gemini 2.5 Flash-Lite}} 
 & \multicolumn{2}{c|}{0-Shot} & 62.92 & 71.32 & 61.32 & 74.48 & 74.04 & 78.88 & 73.04 & 82.92 \\ \cline{2-11}
 & \multirow{5}{*}{4-Shot} 
   & LB    & 65.16 & 77.52 & 64.16 & 83.08 & 82.68 & 89.48 & 81.24 & 91.64 \\
 & & DE    & 65.20 & 79.16 & 65.28 & 82.88 & 80.28 & 88.32 & 79.56 & 90.44 \\
 & & EN    & \textbf{67.36} & \textbf{79.76} & \textbf{67.40} & 83.80 & \textbf{83.68} & \textbf{90.80} & \textbf{83.60} & \textbf{93.48} \\
 & & FR    & 64.44 & 78.40 & 64.36 & \textbf{83.88} & 81.44 & 89.88 & 80.76 & 93.16 \\
 & & MULTI & 64.26 & 78.80 & 63.96 & 82.76 & 82.72 & 90.32 & 82.28 & 92.76 \\ \cline{2-11}
 & \multirow{5}{*}{8-Shot} 
   & LB    & 65.18 & 79.60 & 65.74 & 83.84 & 83.52 & 91.44 & 82.28 & 93.32 \\
 & & DE    & 64.56 & 80.36 & 65.28 & 84.60 & 84.16 & 93.44 & 85.04 & \textbf{95.76} \\
 & & EN    & \textbf{68.52} & \textbf{81.64} & \textbf{69.20} & \textbf{86.36} & \textbf{84.52} & \textbf{94.10} & \textbf{85.92} & 94.88 \\
 & & FR    & 65.90 & 80.40 & 67.68 & 85.12 & 81.16 & 91.04 & 81.60 & 93.36 \\
 & & MULTI & 65.00 & 79.20 & 65.48 & 84.44 & 83.32 & 92.08 & 84.20 & 93.56 \\
\hline

\multirow{11}{*}{\rotatebox[origin=c]{90}{GPT-5 mini}} 
 & \multicolumn{2}{c|}{0-Shot} & 39.32 & 49.88 & 33.04 & 51.04 & 55.96 & 65.24 & 48.80 & 65.72 \\ \cline{2-11}
 & \multirow{5}{*}{4-Shot} 
   & LB    & 38.56 & 49.92 & 32.96 & 51.56 & 61.84 & 71.76 & 54.20 & 73.48 \\
 & & DE    & 37.76 & 50.52 & 33.60 & 53.96 & 61.28 & 72.80 & 53.16 & 73.76 \\
 & & EN    & \textbf{40.60} & \textbf{53.04} & \textbf{35.16} & 54.88 & \textbf{63.60} & \textbf{74.20} & \textbf{56.00} & \textbf{75.28} \\
 & & FR    & 39.16 & 52.52 & 34.80 & \textbf{55.12} & 61.36 & 71.84 & 52.52 & 72.72 \\
 & & MULTI & 38.56 & 50.56 & 32.36 & 52.00 & 63.20 & 72.24 & 53.92 & 73.88 \\ \cline{2-11}
 & \multirow{5}{*}{8-Shot} 
   & LB    & 38.72 & 49.88 & 33.00 & 53.88 & 61.96 & 73.76 & 54.08 & 74.00 \\
 & & DE    & 38.64 & 51.16 & 33.88 & 54.16 & 64.64 & 75.36 & 55.04 & 76.84 \\
 & & EN    & \textbf{41.96} & \textbf{54.28} & \textbf{35.24} & \textbf{56.20} & \textbf{65.48} & \textbf{77.20} & \textbf{58.36} & \textbf{77.56} \\
 & & FR    & 39.72 & 52.36 & 34.48 & 55.24 & 62.68 & 73.36 & 54.84 & 74.68 \\
 & & MULTI & 39.24 & 52.12 & 34.56 & 54.68 & 65.20 & 75.20 & 56.48 & 77.16 \\
\hline
\end{tabular}
\caption{G-Eval results measuring instruction-following performance across Clarity (\textbf{CL}), Coherence (\textbf{CO}), Fluency (\textbf{FL}), and Relevance (\textbf{RE}) in 0-shot, 4-shot, and 8-shot settings, with few-shot examples using different instruction languages (\textbf{IT Lang.}).}
\label{tab:few_shot_results_full}
\end{table*}

\end{document}